%% file: main.tex
\pdfoutput=1

\documentclass[11pt]{article}

\usepackage{ACL2023}

\usepackage{times}
\usepackage{latexsym}

\usepackage[T1]{fontenc}

\usepackage[utf8]{inputenc}

\usepackage{microtype}

\usepackage{inconsolata}

\usepackage{multirow}
\usepackage{tabularx}
\usepackage{amsmath}
\usepackage{amsxtra}
\usepackage{amsfonts}
\usepackage{booktabs}
\usepackage{graphicx}
\usepackage{textcomp}
\usepackage{footnote}
\usepackage{hyperref}
\makesavenoteenv{tabular}
\makesavenoteenv{table}
\usepackage[all]{nowidow}
\usepackage{float} %
\usepackage{refcount}
\usepackage{tablefootnote}
\usepackage{multirow}
\usepackage{siunitx}
\usepackage{enumerate}
\usepackage{stfloats}
\usepackage{footmisc}

\include{small-macros}

\title{Friendly Neighbors: \\Contextualized Sequence-to-Sequence Link Prediction}

\author{Adrian Kochsiek \\
  University of Mannheim \\
  Germany \\
  \texttt{akochsiek} \\
  \texttt{@uni-mannheim.de} \\\And
  Apoorv Saxena \\
  Adobe Research \\
  India \\
  \texttt{apoorvs} \\
  \texttt{@adobe.com} \\\And
  Inderjeet Nair \\
  Adobe Research \\
  India \\
  \texttt{inderjeetnair1}\\
  \texttt{@gmail.com} \\\And
  Rainer Gemulla \\
  University of Mannheim \\
  Germany \\
  \texttt{rgemulla} \\
  \texttt{@uni-mannheim.de}
}

\begin{document}
\maketitle
\begin{abstract}
  We propose \emph{KGT5-context}, a simple sequence-to-sequence model for link
  prediction (LP) in knowledge graphs (KG). Our work expands on KGT5, a recent
  LP model that exploits textual features of the KG, has small model size, and
  is scalable. To reach good predictive performance, however, KGT5 relies on an
  ensemble with a knowledge graph embedding model, which itself is excessively
  large and costly to use. In this short paper, we show empirically that adding
  contextual information---i.e., information about the direct neighborhood of
  the query entity---alleviates the need for a separate KGE model to obtain good
  performance. The resulting KGT5-context model is simple, reduces model size
  significantly, and obtains state-of-the-art performance in our experimental
  study.

\end{abstract}

\section{Introduction}
\label{sec:introduction}

\begin{figure*}[t!]
  \centering
  \includegraphics[width=0.95\textwidth]{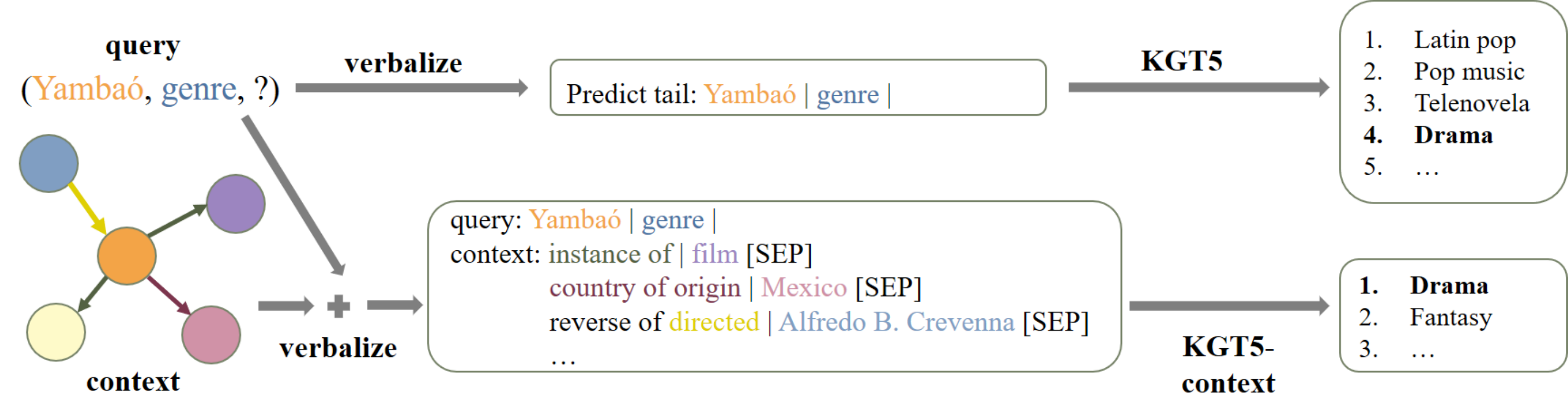}
  \caption{Overview of KGT5-context (at bottom) and comparison to KGT5 (on top);
    real example from Wikidata5M, best viewed in color. KGT5-context differs
    from KGT5 in that it appends the neighboring relations and entities of
    \emph{Yamba\'o} (a drama movie) to the verbalized query. Both models then
    apply T5, sample predictions from the decoder, map the samples to entities,
    and rank by sample logit scores.}
  \label{fig:kgt5-context}
\end{figure*}

A knowledge graph (KG) is a collection of facts describing relations between
real-world entities. Facts are represented in the form of
subject-relation-object ($(s,r,o)$) triples such as (\emph{Brendan Fraser,
  hasWonPrize, Oscar}). In this paper, we study the link prediction (LP)
problem, which is to infer missing links in the KG. We focus on KGs in which the
entities and relations have textual features, such as mention names or
descriptions.

\citet{saxena2022sequence} made a case for large language models (LM) for this
task. They proposed the KGT5 model, which posed the link prediction problem as a
sequence-to-sequence (seq2seq) task. The main advantages of this approach are
that
\begin{enumerate}[(i)]
\item it allows for small model sizes, and
\item it decouples inference cost from the graph size (and, in particular, the number of entities).
\end{enumerate}
They found that KGT5's performance was particularly strong when predicting the
object of new relations for a query entity (e.g., the birthplace of a person),
but fell short of alternative approaches when predicting additional objects for
a known relation (e.g., additional awards won by someone).

To avoid this problem, \citet{saxena2022sequence} used an ensemble of KGT5 with a large knowledge graph embedding (KGE) model (ComplEx~\cite{trouillon2016complex}).
This ensemble did reach good performance but destroyed both advantages (i) and (ii) of using a LM.
In fact, KGE models learn a low-dimensional representation of each entity and each relation in the graph~\cite{bordes2013translating,sun2018rotate,trouillon2016complex}.
Consequently, model size and LP cost are linear in the number of entities in the graph, which can be expensive to use for large-scale KGs.
For example, the currently best-performing model~\cite{cattaneo2022bess} for the large-scale WikiKG90Mv2 benchmark~\cite{hu2021ogblsc} consists of an ensemble of $85$ KGE models;
each taking up more than \qty{86}{GB} of space for parameters.
Though KGE model sizes can be reduced by using compositional embeddings based on text
mentions~\cite{wang2021KEPLER,clouatre2021mlmlm,wang2022simkgc,jiang2023don},
inference cost remains high for large graphs.

We propose and study KGT5-context, which expands on KGT5 by providing contextual
information about the query entity---i.e., information about the direct
neighborhood of the query entity---to facilitate link prediction. Our work is
motivated by the KGE model HittER~\cite{chen2021hitter}, which follows a similar
approach; we use the seq2seq model KGT5 instead of a Transformer-based KGE
model. KGT5-context is very simple:
The only change to KGT5 is that we add a verbalization of the neighborhood of
the query entity to the description of a given LP task; see
Fig.~\ref{fig:kgt5-context} for an example. KGT5-context retains advantages (i)
and (ii) of KGT5.

We performed an experimental study using the Wikidata5M~\cite{wang2021KEPLER}
and WikiKG90Mv2~\cite{hu2021ogblsc} benchmarks. We found that---without further
hyperparameter tuning---KGT5-context reached or exceeded state-of-the-art
performance on both benchmarks using a significantly smaller model size than
alternative approaches. The simple KGT5-context model thus provides a suitable
baseline for further research.

\section{Expanding KGT5 with Context}
\label{sec:kgt5context}

Given a \emph{query} $(s,r,?)$ and a KG, LP is the task to predict new answer entities, i.e., the $?$ slot of the query.
An example is given in Fig.~\ref{fig:kgt5-context}.

\textbf{KGT5}~\cite{saxena2022sequence} treats link prediction as a seq2seq
task. It exploits available textual information for entities and relations, such
as mention names (for both entities and relations) or descriptions.
KGT5's architecture is based on the encoder-decoder Transformer model
T5~\cite{raffel2020exploring}. It uses canonical mentions to verbalize the LP
query to a text sequence of form ``\texttt{predict tail: <subject mention> |
  <relation mention> |~}''. To predict answers, KGT5 samples (exact) candidate
mentions from the decoder; the cost of sampling is independent of the number of
entities in the KG. To train KGT5, \citet{saxena2022sequence}~use standard
training techniques for LLMs: KGT5 is trained on facts in the KG and asked to
generate the true answer using teacher forcing and a cross-entropy loss.

\textbf{KGT5-context} (ours) proceeds in the same way as KGT5 but extends the verbalization of the query.
In particular, we append a textual sequence of the one-hop neighborhood of the query entity $s$ to the verbalized query of KGT5.
As a result, the query entity is contextualized, an approach that has been applied successfully to KGE models before~\cite{chen2021hitter}.
KGT5-context simplifies the prediction problem because additional information that is readily available in the KG is provided along with the query.
In the example of Fig.~\ref{fig:kgt5-context}, the contextual information states that \emph{Yamba\'o} is a Mexican movie.
This information is helpful; e.g., it already rules out the top two predictions of KGT5, which incorrectly suggest that \emph{Yamba\'o} is a piece of music.
For a more detailed analysis, see Sec.~\ref{sec:context}.

\textbf{Verbalization details.} To summarize, we obtain mentions of the entities
and relations in the query as well as in the one-hop neighborhood of the query
entity.
We use these mentions to verbalize the query together with the neighborhood as
``\texttt{query: <query entity mention> | <query relation mention> | context:
  <context relation 1 mention> | <context entity 1 mention> <SEP>
}\ldots''.\footnote{When entity descriptions are available, we include
  ``\texttt{description: <description of query entity>}'' right before the query
  context.} To keep direction of relations, we prepend the relation mention with
``\texttt{reverse of}'' if the query entity acts as an object, i.e., the
relation ``points towards'' the query entity. A real-world example is given in
Fig.~\ref{fig:kgt5-context}. Inspired by neighborhood sampling in
GNNs~\cite{hamilton2017}, we sample up to $k$ (default: $k=100$)
relation-neighbor pairs uniformly, at random, and without replacement.

\section{Experimental Study}
\label{sec:experiments}

We conducted an experimental study to investigate (i) to what extent integrating
context in terms of the entity neighborhood into KGT5 improves link prediction
performance, (ii) whether the use of context can mitigate the necessity for an
ensemble of the text-based KGT5 model with a KGE model, and (iii) for what kind
of queries context is helpful.
We found that:
\begin{enumerate}
\item KGT5-context improved the state-of-the-art performance on Wikidata5M using
  a smaller model (Tab.~\ref{tab:wikidata5m}).
\item KGT5-context was orders of magnitudes smaller than the leading models on WikiKG90Mv2 and reached competitive performance (Tab.~\ref{tab:wikikg90mv2}).%
\item KGT5-context did not benefit further from ensembling with a KGE
  model (Tab.~\ref{tab:query_occurrence}).
\end{enumerate}

\begin{table}
  \centering
  \begin{tabular}{@{}lrrrr@{}}
    \toprule
    \textbf{Dataset} &
                       \multicolumn{1}{r}{\textbf{Entities}} &
                                                               \multicolumn{1}{r}{\textbf{Relations}} &
                                                                                                        \multicolumn{1}{r}{\textbf{Edges}} \\
    \midrule
    Wikidata5M          & 4.8M & 828   & 21M  \\
    WikiKG90Mv2         & 91M  & 1,387 & 601M \\
    \bottomrule
  \end{tabular}%
  \caption{
    Dataset statistics.
  }
  \label{tab:dataset_statistics}
\end{table}

{
\aboverulesep = 0.20ex
\belowrulesep = 0.30ex
\begin{table*}[ht!]
  \centering
\resizebox{2.\columnwidth}{!}{%
  \begin{tabular}{@{}lrrrrrcc@{}}
    \toprule
    & & & & &
    & \multirow{2}{*}{\begin{tabular}[l]{@{}c@{}}\textbf{Add.~requirements}\\\textbf{for inference}\end{tabular}}
    & \multirow{2}{*}{\begin{tabular}[l]{@{}c@{}}\textbf{Pre-}\\\textbf{trained}\end{tabular}} \\
    \textbf{Model} & \textbf{MRR}   & \textbf{Hits@1} & \textbf{Hits@3} & \textbf{Hits@10} & \textbf{Params} & \\ \midrule
    SimplE $^{\dagger}$         & 0.296          & 0.252  & 0.317           & 0.377            & 2,400M & - & no           \\
    ComplEx $^{\dagger\dagger}$       & \underline{0.308} & \underline{0.255}           & -  & \underline{0.398}   & \underline{614M} & - & no            \\ \midrule
    SimKGC & 0.212 & 0.182 & 0.223 & 0.266 & 220M & entity embeddings & yes\\
    KGT5 $^{\ddagger}$          & 0.300 & 0.267   & 0.318  & 0.365   & \underline{60M} & - & no   \\
    \hspace{2mm} + ComplEx $^{\ddagger}$ & 0.336 & 0.286  & 0.362  & 0.426 & 674M & - & no           \\
    KGT5-context (ours) & \underline{0.378} & \underline{0.350} & \underline{0.396} & \underline{0.427} & \underline{60M} & 1-hop neighborhood & no \\
    \midrule
    SimKGC + Desc. $^{\ddagger\ddagger}$ & 0.358 & 0.313 & 0.376 & 0.441 & 220M & entity embeddings & yes \\
    \hspace{2mm} + Hard Negative Ensemble $^\$$ & 0.420 & 0.381 & 0.435 & \underline{\textbf{0.490}} & 1,100M & entity embeddings & yes \\
    KGT5 + Desc. & 0.381 & 0.357 & 0.397 & 0.422 & \underline{60M} & - & no \\
    KGT5-context + Desc. (ours) & \underline{\textbf{0.426}} & \underline{\textbf{0.406}} & \underline{\textbf{0.440}} & 0.460 & \underline{60M} & 1-hop neighborhood & no \\ \bottomrule
  \end{tabular}%
}
  \caption{
    Link prediction results on Wikidata5M,  test split. The first group does not make use of textual information, the second group uses mention names, the third group additionally entity descriptions. Best per group underlined, best overall bold.
    Marked results are from
    $\dagger$
    \citet{zhu2019graphvite},
    $\dagger\dagger$~\citet{kochsiek2021parallel}, $\ddagger$~\citet{saxena2022sequence},
    $\ddagger\ddagger$~\citet{wang2022simkgc}, $\$$~\citet{jiang2023don}. Additional results in
    Tab.~\ref{tab:wikidata5m_full} (appendix).}
  \label{tab:wikidata5m}
\end{table*}
}

\subsection{Experimental Setup}
\label{sec:setup}

Source code and configuration are available at \url{https://github.com/uma-pi1/kgt5-context}.

\textbf{Datasets.}
\label{sec:datasets}
We evaluate KGT5-context on two commonly used large-scale link prediction
benchmarks. Wikidata5M~\cite{wang2021KEPLER} is the induced graph of the 5M
most-frequent entities of the Wikidata KG.
WikiKG90Mv2~\cite{hu2021ogblsc} contains more than 90M entities and over 600M facts.
In contrast to Wikidata5M, it is only evaluated on tail prediction, i.e.,
$(s,r,?)$ queries. Dataset statistics are summarized in
Tab.~\ref{tab:dataset_statistics}. For Wikidata5M and
WikiKG90Mv2\footnote{\label{fn:wikidata90m}We directly used mentions of entities
  and relations for WikiKG90Mv2, instead of the textual embeddings used by other
  models. For this reason, the benchmark authors~\cite{hu2021ogblsc} did not
  provide us with scores on the hidden test set. The mentions used to be
  provided with the dataset but have been removed by now; we obtained them from
  \url{https: //github.com/apoorvumang/kgt5.}}, we used the entity mentions
provided on the KGT5 webpage.
For Wikidata5M, we also consider the usefulness of entity descriptions, which are provided with the dataset and have been used in some prior studies~\cite{wang2022simkgc,jiang2023don}.
Note that we do not use these descriptions by default, and clearly mark throughout when they have been used.

\textbf{Metrics.} We follow the standard procedure to evaluate model quality for the link prediction task.
In particular, for each test triple $(s,r,o)$, we rank all triples of the form $(s,r,?)$ (and $(?,r,o)$ on Wikidata5M) by their predicted scores.
For KGT5 and KGT5-context, we instead sample from the decoder and ignore outputs that do not correspond to an existing entity mention.
For all models, we filter out all true answers other than the test triple that occur either in the train, valid or test data.
Finally, we determine the mean reciprocal rank (MRR) and Hits@K over all test triples.
In case of ties, we use the mean rank to avoid misleading results~\cite{sun2020re}.

\textbf{Settings.} We mainly follow the setting of KGT5. For all experiments, we
used the same T5 architecture (T5-small for Wikidata5M, T5-base for WikiKG90Mv2)
without any pretrained weights. Training from scratch ensures test data is
unseen during (pre-)training and avoids
leakage. %
We used the SentencePiece tokenizer pretrained by~\cite{raffel2020exploring}. We
trained on 8 A100-GPUs with a batch size of $32$ (effective batch size of $256$)
using the AdaFactor optimizer. No dataset-specific hyperparameter optimization
was performed. For KGT5-context, we sampled up to $100$ neighbors per query
entity or up to an input sequence length of 512 tokens. For inference, we
obtained $500$ samples from the decoder.

\textbf{Models.} On Wikidata5M, we compare KGT5-context to the KGE models
ComplEx~\cite{trouillon2016complex} and SimplE~\cite{kazemi2018simple} (only
graph structure used), the compositional KGE model SimKGC~\cite{wang2022simkgc},
its extension utilizing hard negatives~\cite{jiang2023don}, and the seq2seq
model KGT5~\cite{saxena2022sequence}.
The model of \citet{jiang2023don} is an ensemble of multiple SimKGC models, each trained with a different strategy for selecting negatives.
Note that in contrast to KGT5 and KGT5-context, SimKGC is based on pretrained models (BERT transformers).
During prediction, all text-based models require access to entity and relation mentions and, when used, the description of the query entity.
SimKGC additionally requires access to precomputed entity embeddings, KGT5-context to the 1-hop neighborhood of the query entity in the KG.

On WikiKG90M, we compare to the models presented on the official leaderboard.\footref{fn:wikidata90m}
Here, the best-performing approaches are large ensembles of multiple KGE models.

\subsection{Link Prediction Performance}
\label{sec:results}

Link prediction performance on Wikidata5M is shown in Tab.~\ref{tab:wikidata5m};
additional baselines are given in Tab.~\ref{tab:wikidata5m_full} (appendix).
Generally, we found that textual information was highly beneficial. KGT5-context
was the only model that improved upon KGE models (which do not use textual
information) when only mention information was available. Moreover, KGT5-context
obtained better predictive performance than the ensemble of KGT5 with the
ComplEx KGE model. Entity descriptions provided further improvements; they hold
valuable information for this benchmark. With these descriptions, KGT5-context
outperformed traditional KGE models by up to 12pp in terms of MRR, with a model
size reduction of 90-98\%. Likewise, KGT5-context improved on KGT5 by 12pp, on
the KGT5+Complex ensemble by almost 9pp, and performed roughly on-par with the
current state-of-the-art SimKGC ensemble model, which is significantly larger.

The results on the much larger WikiKG90Mv2 are shown in
Tab.~\ref{tab:wikikg90mv2}.\footref{fn:wikidata90m} Here, KGT5-context is multiple orders of magnitude
smaller than the currently best-performing models,\footnote{\label{ftn:bess}The
  parameter count in Tab.~\ref{tab:wikikg90mv2} corresponds to the size of the
  largest model in an ensemble, not the overall model size. For example, BESS~\cite{cattaneo2022bess}
  consists of 85 models and the complete ensemble has 2.6T parameters; the
  KGT5-context model is 5 orders of magnitude smaller.} and improves validation
MRR by almost 1pp. %

{
\aboverulesep = 0.15ex
\belowrulesep = 0.25ex
\begin{table}[t!]
\centering
\resizebox{\columnwidth}{!}{%
\begin{tabular}{@{}lrrr@{}}
\toprule
\multicolumn{1}{c}{\textbf{Model}} &
  \multicolumn{1}{c}{\textbf{\begin{tabular}[c]{@{}c@{}}Test\\ MRR\end{tabular}}} &
  \multicolumn{1}{c}{\textbf{\begin{tabular}[c]{@{}c@{}}Valid\\ MRR\end{tabular}}} &
  \multicolumn{1}{c}{\textbf{Params}} \\ \midrule
ComplEx & 0.141          & 0.182          & 18.2B \\
TransE  & 0.082          & 0.110          & 18.2B \\
ComplEx-Concat  & 0.176 & 0.205          & 18.2B \\
TransE-Concat   & 0.176 & 0.206          & 18.2B \\
PIE-RM  & 0.212 & 0.254 & 18.2B\footref{ftn:bess} \\
DGLKE + Rule Mining & 0.249 & 0.292 & 18.2B\footref{ftn:bess} \\
BESS & \textbf{0.254} & 0.292 & 23.3B\footref{ftn:bess} \\
KGT5, T5 small\footref{fn:wikidata90m}       & -              & 0.221 & 60M   \\
  KGT5-context, T5 base (ours)\footref{fn:wikidata90m}             & -  & \textbf{0.301} & 220M   \\ \bottomrule
\end{tabular}%
}
\caption{
  Link prediction results on WikiKG90Mv2. Baseline numbers are from the official leaderboard of OGB-LSC~\cite{hu2021ogblsc}.
}
\label{tab:wikikg90mv2}
\end{table}
}

\subsection{Analysis}
\label{sec:context}

\begin{table}[t!]
  \centering
  \resizebox{\columnwidth}{!}{%
    \begin{tabular}{lrrrrr}
      \toprule
      \textbf{Model} &
      \textbf{0} &
      \textbf{1-10} &
      \textbf{>10} &
      \textbf{All} \\
      \midrule
      ComplEx & 0.534 & 0.351 & \textbf{0.045} & 0.296 \\
      KGT5 & 0.624 & 0.215 & 0.015 & 0.300 \\
      KGT5-context (ours) & \textbf{0.738} & \textbf{0.415} & 0.014 & 0.378 \\
      \midrule
      KGT5 + ComplEx & 0.624 & 0.351 & 0.045 & 0.336 \\
      KGT5-context + ComplEx & \textbf{0.738} & 0.351 &\textbf{0.045} & \textbf{0.379} \\
      \bottomrule
    \end{tabular}
  }
  \caption{Test MRR on Wikidata5M grouped by query frequency during training.}
  \label{tab:query_occurrence}
\end{table}

To investigate in which cases context information was beneficial, we empirically
analyzed LP performance w.r.t.~(i) query frequency and (ii) the degree of the
query entity. We also sampled predictions and summarize our general
observations.

\textbf{Query frequency.} The \emph{frequency} of a test query $(s,r,?)$ is the
number of answers to the query already available in the training data. For
example, queries for N:1 relations have frequency 0, whereas queries for 1:N
relations can have large frequency for high-degree query entities. We bucketized
the test queries of Wikidata5M into low, medium, and high frequency queries and
report average MRR for various models in Tab.~\ref{tab:query_occurrence}.
Generally, high-frequency queries appear harder to answer. These queries have
many known true answers already (tying up model capacity); there may be many
additional, potentially unrelated answers and incompleteness of the KG may be a
concern during evaluation.
In contrast, a low-frequency query such as \emph{(Brendan Fraser, instance Of,
  ?)} has few or no known answer and might be easier to infer, even when the
combination of this particular subject and relation was not yet seen during
training.

\textbf{Ensemble with KGE models.} In general, the prior KGT5 model performed
reasonably well on queries that did not occur in the training data, but was
outperformed by a large amount by ComplEx on queries seen multiple times. Hence,
both models complemented each other in an ensemble. KGT5-context strongly
improved performance over ComplEx, KGT5, and the KGT5+Complex ensemble for low-
and medium-frequency queries. For this reason, an ensemble between KGT5-context
and ComplEx only brought negligible benefits, but has substantial drawbacks.
Consequently, an ensemble of KGT5-context with a KGE model is not needed and
should not be used.

\textbf{Entity degree.} We also investigated the benefit of contextual
information w.r.t.\ to the degree of the query entity (see
Fig.~\ref{fig:mrr_by_degree} in the appendix). We found that KGT5-context was
beneficial and performed well on query entities with a degree of up to 100. For
entities with very large degrees (i.e., nodes with more than 100 or even 1000s
of neighbors), ComplEx showed benefits. As before, we feel that these
performance benefits are negligible considering the increase in model size and
decrease in scalability.

\textbf{Anecdotal results.} We manually probed some predictions of KGT5-context
and found the context is especially beneficial when (i) the entity mention only
provides limited information about the entity, and/or when (ii) the answer to
the query is contained in the one-hop neighborhood.

A case of (i) is shown in Fig.~\ref{fig:kgt5-context}, a real example. Here,
KGT5 was able to capture the geographic region of the real-world entity only
based on its mention. Based on this geographic notion, it proposed the music
genre \emph{Latin pop} but was unaware that the entity is a movie. This useful
information can be obtained directly from the one-hop neighborhood and, indeed,
was exploited by KGT5-context.

For Wikidata5M, the correct answer entity appears in the one-hop neighborhood of
the query entity for about $7\%$ of the validation triples. But even when the
answer does not directly appear in the context, it may contain entities strongly
hinting at the correct answer. For example, it is easier to predict that an
entity has occupation \emph{biochemist}, when the context already contains the
information that the entity is a \emph{chemist}.

\section{Conclusion}
\label{sec:conclusion}

We proposed and studied KGT5-context, a sequence-to-sequence model for link
prediction in knowledge graphs. KGT5-context extends the KGT5 model of
\citet{saxena2022sequence} by using contextual information of the query entity
for prediction. KGT5-context is simple, small, and scalable, and it obtained or
exceeded state-of-the-art performance in our experimental study. It thus
provides a suitable baseline for further research in this area. A natural
direction, for example, is to explore approaches that integrate contextual
information in a less naive way than KGT5-context does.

\section*{Limitations}

KGT5-context relies on the textual mentions of entities and relations (and,
optionally, entity descriptions). Therefore, it is only applicable to KGs that
provide such information. KGT5-context may be able to handle some entities
without textual features when well-described by their neighborhood; we did not
investigate this though.

To use KGT5-context for prediction, the KG has to be queried to obtain
context information, i.e., the one-hop neighborhood of the query entity.
KGT5-context thus cannot be used without the underlying KG.

The verbalized neighborhood of the query entity leads to long input sequences,
which in turn may induce higher memory consumption and higher computational
cost during training. Overall, training KGT5-context is typically more expensive
than training traditional KGE models, which can be
tuned~\cite{kochsiek2023start} and trained
efficiently~\cite{lerer2019pytorch,kochsiek2021parallel,zheng2020dgl}.

For inference, KGT5-context first samples relation-neighbor pairs for
contextualization, and then samples possible answers from the decoder. These
sampling steps can lead to variance in predictive performance. We found this
effect to be negligible on Wikidata5M, but it may be larger on other datasets.

\clearpage
\section*{Ethics Statement}

This research uses publicly available data and benchmarks for evaluation.
We believe that this research was conducted in an ethical manner and in compliance with all relevant laws and regulations.

\bibliography{main}
\bibliographystyle{acl_natbib}

\clearpage

\renewcommand{\bottomfraction}{.9}%
\appendix

\begin{minipage}{15cm}
\section*{Appendix}
\label{sec:appendix}
\end{minipage}

\begin{figure*}[b]
  \centering
  \includegraphics[width=1.1\columnwidth]{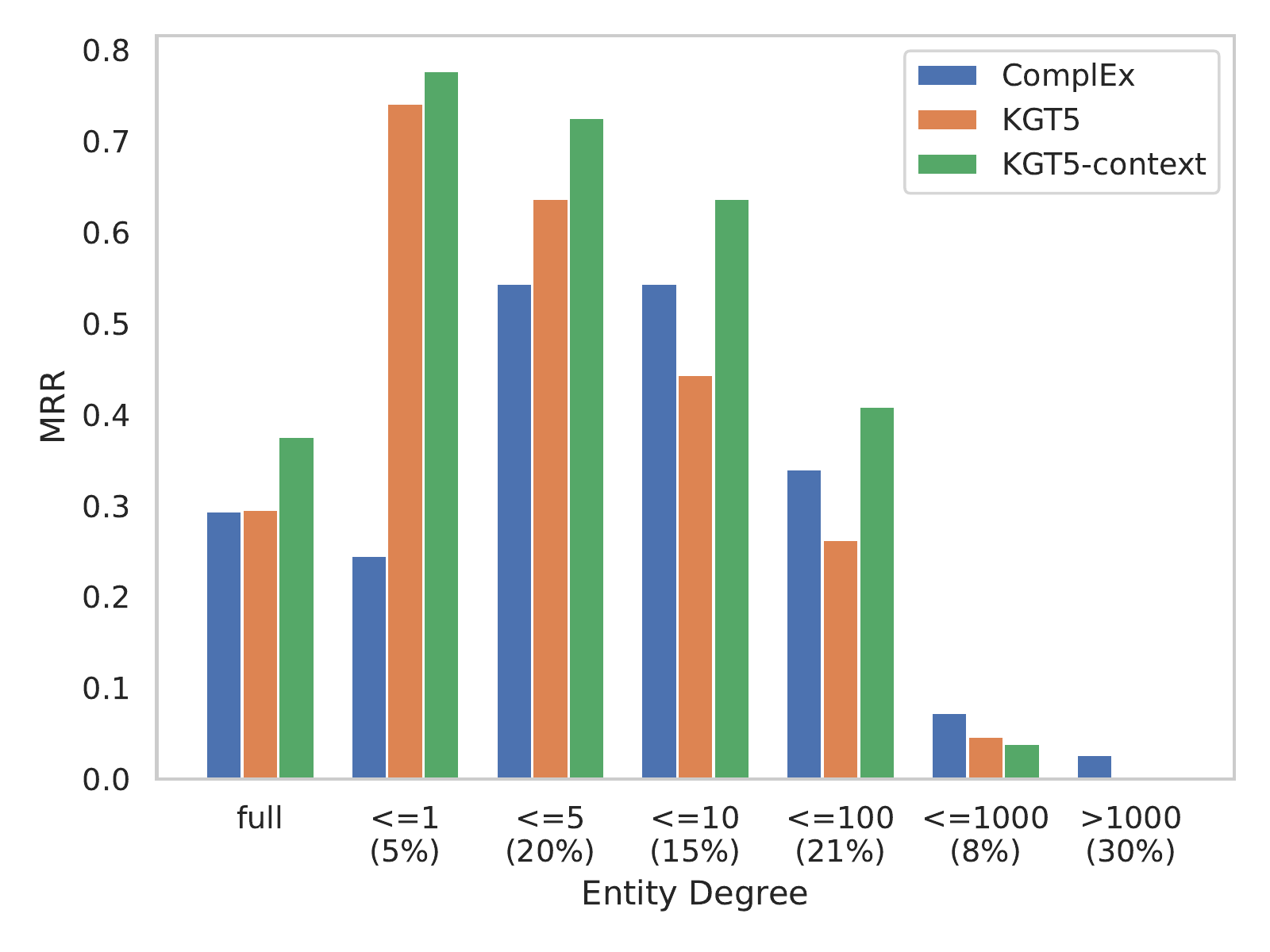}
  \caption{MRR grouped by entity degree on Wikidata5M. Group weight is given in brackets.}
  \label{fig:mrr_by_degree}
\end{figure*}

\begin{table*}[b]
  \centering

  \begin{tabular}{@{}lrrrrr@{}}
    \toprule
    \textbf{Model} & \textbf{MRR}   & \textbf{Hits@1} & \textbf{Hits@3} & \textbf{Hits@10} & \textbf{Params} \\ \midrule
    TransE~\cite{bordes2013translating} $^\dagger$        & 0.253          & 0.170            & 0.311           & 0.392            & 2,400M            \\
    DistMult~\cite{yang2015embedding} $^\dagger$      & 0.253          & 0.209           & 0.278           & 0.334            & 2,400M            \\
    RotatE~\cite{sun2018rotate} $^\dagger$         & 0.290           & 0.234           & 0.322           & 0.390             & 2,400M            \\
    DKRL~\cite{Xie_Liu_Jia_Luan_Sun_2016} $^{\$}$         & 0.160           & 0.120           & 0.181           & 0.229            & 20M            \\
    KEPLER~\cite{wang2021KEPLER} $^{\$}$         & 0.210           & 0.173           & 0.224           & 0.277            & 125M            \\
    MLMLM~\cite{clouatre2021mlmlm} $^{\dagger\dagger}$    &   0.223   &   0.201   &   0.232   &   0.264   &   355M   \\
    \bottomrule
  \end{tabular}%

  \caption{
    Additional link prediction results on Wikidata5M from prior work.
    Results are from $\dagger$ \citet{zhu2019graphvite}.
    $\$$ \citet{wang2021KEPLER}.
    $\dagger\dagger$ \citet{clouatre2021mlmlm}.
  }
  \label{tab:wikidata5m_full}
\end{table*}

\end{document}

%% file: small-macros.tex
\newcommand{\eat}[1]{} %

\newcommand{\td}[2]{\if*#1\else^{#1}\fi\if*#2\else_{#2}\fi} %

\newcommand\join\Join %

\DeclareSymbolFont{txsymbolsC}{U}{txsyc}{m}{n}
\SetSymbolFont{txsymbolsC}{bold}{U}{txsyc}{bx}{n}
\DeclareFontSubstitution{U}{txsyc}{m}{n}
\DeclareMathSymbol{\ljoin}{\mathrel}{txsymbolsC}{88}
\DeclareMathSymbol{\rjoin}{\mathrel}{txsymbolsC}{89}

\newsavebox\setminusbox
\newlength\setminuslen

\newcolumntype{C}{>{$\displaystyle}c<{$}} %
\newcolumntype{L}{>{$\displaystyle}l<{$}} %
\newcolumntype{R}{>{$\displaystyle}r<{$}} %
\newcolumntype{H}{>{\setbox0=\hbox\bgroup}c<{\egroup}@{}} %

\newcommand{\B}[3]{B\if*#1\else_{#1}\fi(#2,#3)} %
\newcommand{\I}[3]{I\if*#1\else_{#1}\fi(#2,#3)} %

\makeatletter
\def\imod#1{\allowbreak\mkern10mu({\operator@font mod}\,\,#1)}
\makeatother

\newlength\hspaceoflen

\newcommand\vect[1]{{\boldsymbol{#1}}}
\newcommand\va{\vect{a}}
\newcommand\vb{\vect{b}}
\newcommand\vc{\vect{c}}
\newcommand\vd{\vect{d}}
\newcommand\ve{\vect{e}}
\newcommand\vf{\vect{f}}
\newcommand\vg{\vect{g}}
\newcommand\vh{\vect{h}}
\newcommand\vi{\vect{i}}
\newcommand\vj{\vect{j}}
\newcommand\vk{\vect{k}}
\newcommand\vl{\vect{l}}
\newcommand\vm{\vect{m}}
\newcommand\vn{\vect{n}}
\newcommand\vo{\vect{o}}
\newcommand\vp{\vect{p}}
\newcommand\vq{\vect{q}}
\newcommand\vr{\vect{r}}
\newcommand\vs{\vect{s}}
\newcommand\vt{\vect{t}}
\newcommand\vu{\vect{u}}
\newcommand\vv{\vect{v}}
\newcommand\vw{\vect{w}}
\newcommand\vx{\vect{x}}
\newcommand\vy{\vect{y}}
\newcommand\vz{\vect{z}}

\newcommand\mA{\vect{A}}
\newcommand\mB{\vect{B}}
\newcommand\mC{\vect{C}}
\newcommand\mD{\vect{D}}
\newcommand\mE{\vect{E}}
\newcommand\mF{\vect{F}}
\newcommand\mG{\vect{G}}
\newcommand\mH{\vect{H}}
\newcommand\mI{\vect{I}}
\newcommand\mJ{\vect{J}}
\newcommand\mK{\vect{K}}
\newcommand\mL{\vect{L}}
\newcommand\mM{\vect{M}}
\newcommand\mN{\vect{N}}
\newcommand\mO{\vect{O}}
\newcommand\mP{\vect{P}}
\newcommand\mQ{\vect{Q}}
\newcommand\mR{\vect{R}}
\newcommand\mS{\vect{S}}
\newcommand\mT{\vect{T}}
\newcommand\mU{\vect{U}}
\newcommand\mV{\vect{V}}
\newcommand\mW{\vect{W}}
\newcommand\mX{\vect{X}}
\newcommand\mY{\vect{Y}}
\newcommand\mZ{\vect{Z}}

\DeclareMathAlphabet{\mathcal}{OMS}{cmsy}{m}{n}

\DeclareMathAlphabet\mathbfcal{OMS}{cmsy}{b}{n}

\accentedsymbol\Abar{{\bar A}}
\accentedsymbol\Bbar{{\bar B}}
\accentedsymbol\Cbar{{\bar C}}
\accentedsymbol\Dbar{{\bar D}}
\accentedsymbol\Ebar{{\bar E}}
\accentedsymbol\Fbar{{\bar F}}
\accentedsymbol\Gbar{{\bar G}}
\accentedsymbol\Hbar{{\bar H}}
\accentedsymbol\Ibar{{\bar I}}
\accentedsymbol\Jbar{{\bar J}}
\accentedsymbol\Kbar{{\bar K}}
\accentedsymbol\Lbar{{\bar L}}
\accentedsymbol\Mbar{{\bar M}}
\accentedsymbol\Nbar{{\bar N}}
\accentedsymbol\Obar{{\bar O}}
\accentedsymbol\Pbar{{\bar P}}
\accentedsymbol\Qbar{{\bar Q}}
\accentedsymbol\Rbar{{\bar R}}
\accentedsymbol\Sbar{{\bar S}}
\accentedsymbol\Tbar{{\bar T}}
\accentedsymbol\Ubar{{\bar U}}
\accentedsymbol\Vbar{{\bar V}}
\accentedsymbol\Wbar{{\bar W}}
\accentedsymbol\Xbar{{\bar X}}
\accentedsymbol\Ybar{{\bar Y}}
\accentedsymbol\Zbar{{\bar Z}}

\accentedsymbol\abar{{\bar a}}
\accentedsymbol\bbar{{\bar b}}
\accentedsymbol\cbar{{\bar c}}
\accentedsymbol\dbar{{\bar d}}
\accentedsymbol\ebar{{\bar e}}
\accentedsymbol\fbar{{\bar f}}
\accentedsymbol\gbar{{\bar g}}
\makeatletter
\@ifundefined{hbar}{}{
        \let\hbar\@undefined
}
\makeatother
\accentedsymbol\hbar{{\bar h}}
\accentedsymbol\ibar{{\bar i}}
\accentedsymbol\jbar{{\bar j}}
\accentedsymbol\kbar{{\bar k}}
\accentedsymbol\lbar{{\bar l}}
\accentedsymbol\mbar{{\bar m}}
\accentedsymbol\nbar{{\bar n}}
\makeatletter
\@ifundefined{obar}{}{
        \let\obar\@undefined
}
\makeatother
\accentedsymbol{\obar}{{\bar o}}
\accentedsymbol\pbar{{\bar p}}
\accentedsymbol\qbar{{\bar q}}
\accentedsymbol\rbar{{\bar r}}
\accentedsymbol\sbar{{\bar s}}
\accentedsymbol\tbar{{\bar t}}
\accentedsymbol\ubar{{\bar u}}
\accentedsymbol\vbar{{\bar v}}
\accentedsymbol\wbar{{\bar w}}
\accentedsymbol\xbar{{\bar x}}
\accentedsymbol\ybar{{\bar y}}
\accentedsymbol\zbar{{\bar z}}

\accentedsymbol\mAhat{{\hat\mA}}
\accentedsymbol\mBhat{{\hat\mB}}
\accentedsymbol\mChat{{\hat\mC}}
\accentedsymbol\mDhat{{\hat\mD}}
\accentedsymbol\mEhat{{\hat\mE}}
\accentedsymbol\mFhat{{\hat\mF}}
\accentedsymbol\mGhat{{\hat\mG}}
\accentedsymbol\mHhat{{\hat\mH}}
\accentedsymbol\mIhat{{\hat\mI}}
\accentedsymbol\mJhat{{\hat\mJ}}
\accentedsymbol\mKhat{{\hat\mK}}
\accentedsymbol\mLhat{{\hat\mL}}
\accentedsymbol\mMhat{{\hat\mM}}
\accentedsymbol\mNhat{{\hat\mN}}
\accentedsymbol\mOhat{{\hat\mO}}
\accentedsymbol\mPhat{{\hat\mP}}
\accentedsymbol\mQhat{{\hat\mQ}}
\accentedsymbol\mRhat{{\hat\mR}}
\accentedsymbol\mShat{{\hat\mS}}
\accentedsymbol\mThat{{\hat\mT}}
\accentedsymbol\mUhat{{\hat\mU}}
\accentedsymbol\mVhat{{\hat\mV}}
\accentedsymbol\mWhat{{\hat\mW}}
\accentedsymbol\mXhat{{\hat\mX}}
\accentedsymbol\mYhat{{\hat\mY}}
\accentedsymbol\mZhat{{\hat\mZ}}

\accentedsymbol\vahat{{\hat\va}}
\accentedsymbol\vbhat{{\hat\vb}}
\accentedsymbol\vchat{{\hat\vc}}
\accentedsymbol\vdhat{{\hat\vd}}
\accentedsymbol\vehat{{\hat\ve}}
\accentedsymbol\vfhat{{\hat\vf}}
\accentedsymbol\vghat{{\hat\vg}}
\accentedsymbol\vhhat{{\hat\vh}}
\accentedsymbol\vihat{{\hat\vi}}
\accentedsymbol\vjhat{{\hat\vj}}
\accentedsymbol\vkhat{{\hat\vk}}
\accentedsymbol\vlhat{{\hat\vl}}
\accentedsymbol\vmhat{{\hat\vm}}
\accentedsymbol\vnhat{{\hat\vn}}
\accentedsymbol\vohat{{\hat\vo}}
\accentedsymbol\vphat{{\hat\vp}}
\accentedsymbol\vqhat{{\hat\vq}}
\accentedsymbol\vrhat{{\hat\vr}}
\accentedsymbol\vshat{{\hat\vs}}
\accentedsymbol\vthat{{\hat\vt}}
\accentedsymbol\vuhat{{\hat\vu}}
\accentedsymbol\vvhat{{\hat\vv}}
\accentedsymbol\vwhat{{\hat\vw}}
\accentedsymbol\vxhat{{\hat\vx}}
\accentedsymbol\vyhat{{\hat\vy}}
\accentedsymbol\vzhat{{\hat\vz}}